\documentclass[letterpaper, 10 pt, conference]{ieeeconf}  

\IEEEoverridecommandlockouts                              

\usepackage{subfiles}
\usepackage{amsmath}
\usepackage{amsfonts}
\usepackage{amsthm,mathrsfs}
\usepackage{bm}
\usepackage{cases}
\usepackage{balance}
\usepackage{mathtools}
\usepackage[urlcolor=blue]{hyperref} 
\usepackage{cleveref}
\usepackage{xcolor}
\usepackage{algorithmic,algorithm}
\usepackage{shortcuts}

\usepackage{tabularx,multirow}

\title{\LARGE \bf
Hybrid Data-Driven Predictive Control for Robust and Reactive Exoskeleton Locomotion Synthesis
}

\author{Kejun Li$^{1}$, Jeeseop Kim$^{2}$, Maxime Brunet$^{3}$, Marine P\'etriaux$^{3}$, Yisong Yue$^{4}$, Aaron D. Ames$^{2,5}$
\thanks{This research was supported by Wandercraft. Research involving human subjects was conducted under IRB No. 21-0693.}
\thanks{$^{1}$K. Li is with the Department of Computation and Neural Systems, Caltech, Pasadena, CA 91125, USA, {\tt\small kli5@caltech.edu}.}\newline
\thanks{$^{2}$J. Kim and A. D. Ames are with the Department of Mechanical and Civil Engineering, Caltech, Pasadena, CA 91125, USA, {\tt\small \{jeeseop, ames\}@caltech.edu}.}\newline
\thanks{$^{3}$J. Brunet and M. P\'etriaux are with Wandercraft, Paris, France, {\tt\small \{maxime.brunet,marine.petriaux\}@wandercraft.eu}.}\newline
\thanks{$^{4}$Y. Yue is with the Department of Computing and Mathematical Sciences, Caltech, Pasadena, CA 91125, USA, {\tt\small yyue@caltech.edu}.}%
\thanks{$^{5}$ A. D. Ames is with the Department of Control and Dynamical Systems, Caltech, Pasadena, CA 91125, USA, {\tt\small ames@caltech.edu}.}%
}

\begin{document}

\maketitle
\thispagestyle{empty}
\pagestyle{empty}


\begin{abstract} 
Robust bipedal locomotion in exoskeletons requires the ability to dynamically react to changes in the environment in real time. This paper introduces the \emph{hybrid} data-driven predictive control (HDDPC) framework, an extension of the data-enabled predictive control, that addresses these challenges by simultaneously planning foot contact schedules and continuous domain trajectories. The proposed framework utilizes a Hankel matrix-based representation to model system dynamics, incorporating step-to-step (S2S) transitions to enhance adaptability in dynamic environments. By integrating contact scheduling with trajectory planning, the framework offers an efficient, unified solution for locomotion motion synthesis that enables robust and reactive walking through online replanning. We validate the approach on the Atalante exoskeleton, demonstrating improved robustness and adaptability.
\end{abstract}

\section{Introduction} 
Recent advancements in humanoid and bipedal robotics have shown significant potential for meaningful contributions to human society, especially given their successful translation to exoskeletons \cite{harib2018feedback,gurriet2018towards,kerdraon2021evaluation}. However, robotic bipedal movement remains inherently complex, characterized by high degrees of freedom (DoFs) and underactuated hybrid systems \cite{grizzle2014models}. These complexities, combined with the challenges of accurately modeling such systems, are further compounded in lower-body exoskeletons by the unpredictable dynamics of user-exoskeleton interactions.

Traditional model-based approaches to locomotion synthesis often involve offline trajectory design, as seen in \cite{reher2020algorithmic,westervelt2003hybrid}. However, these methods can lead to large-scale optimization problems and are prone to convergence issues. Alternatively, strategies like Raibert-style foot placement \cite{reher2021dynamic}, or reduced-order models like linear inverted pendulum and its variants \cite{kajita2003biped,wensing2013high,englsberger2015three} focus on online stabilization via replanning. While computationally efficient, these methods rely on simplifying assumptions, which can limit their effectiveness when there are substantial model mismatches. 

On the other end of the spectrum, pure data-driven approaches like reinforcement learning (RL) have made significant advances \cite{hoeller2024anymal,li2024reinforcement}, employing techniques like domain randomization to improve generalization from simulations to real-world conditions. However, RL poses unique challenges for exoskeletons, which are more kinematically constrained than humanoid robots, as they must align with users' movement capabilities and often operate near singularities to achieve anthropomorphic behavior. These requirements may restrict the policy search space, making it harder for RL to efficiently explore viable solutions and resulting a long training time and more demanding reward shaping process to generate a reasonable policy.

\begin{figure}[t!]
\centering
\includegraphics[draft=false, width=\linewidth]{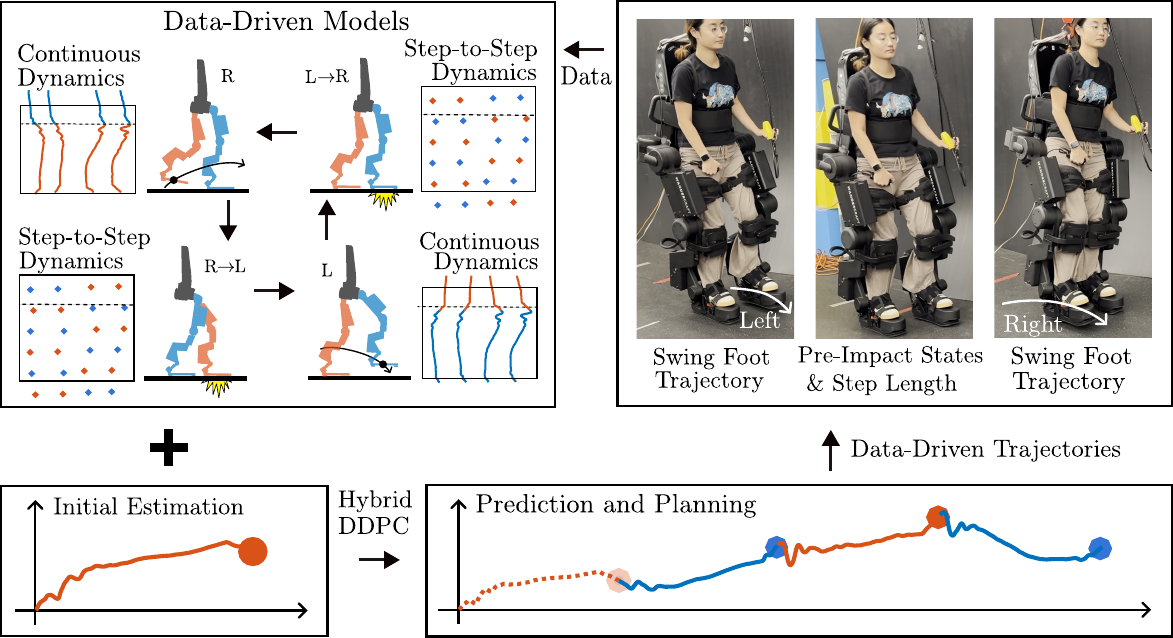}
\caption{Overview of Hybrid DDPC}
\label{fig:overview}
\vspace{-2.0em}
\end{figure}
Data-driven approaches, grounded in behavioral systems theory \cite{markovsky2006exact}, provide a middle ground between model-based methods and learning-based approaches. These methods effectively learn linear time-invariant (LTI) system models, and have been successfully integrated into predictive control frameworks, such as data-enabled predictive control (DeePC) or data-driven predictive control (DDPC) \cite{coulson2019data}. 
DDPC is computationally efficient, and has been successfully applied to systems like quadrupeds, interconnected systems, and assistive devices \cite{fawcett2022toward,fawcett2023distributed,ulkir2021data}. 
Its ability to design policies based on past feasible trajectories makes it a more suitable choice for real-time control in environments with constrained kinematic spaces, where rapid and reliable control is critical.

To achieve robust walking on exoskeletons, it is necessary to generate both stable periodic motions while replanning foot placement online to account for disturbances and locomote in constrained environments. To reduce computational complexity in locomotion motion planning, a common strategy is to separate footstep planning (the discrete dynamics) from motion synthesis (the continuous swing phase dynamics), often relying on heuristics to plan contacts with simplified models \cite{stephens2010push}. 
Recent work has effectively utilized reduced-order models to plan foot placement through the step-to-step (S2S) dynamics \cite{xiong20223}, with the result being robust and agile bipedal locomotion. 
However, this separation requires the reduced order models be valid (e.g., the robot must have light legs), and can lead to instability in dynamic or challenging tasks. This makes it necessary to optimize both the contact schedule and motion planning together \cite{winkler2018gait} or learn the contact schedule through data-driven methods \cite{seyde2019locomotion}. Missing from the existing work is a means to combine the discrete step-to-step planning with continuous trajectory generation in a data-driven manner.

\begin{figure}
  \centering
  \noindent\includegraphics[width=\linewidth]{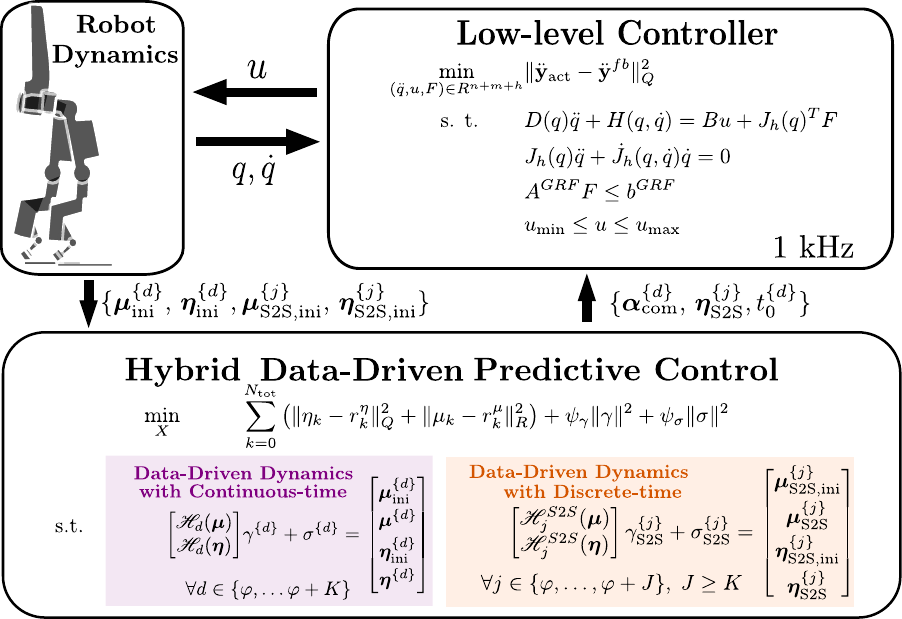}
  \caption{Control Overview for Hybrid Data-Driven Predictive Control (HDDPC). We utlilize a layered architecture with HDDPC planner and low-level controller}
  \label{fig:HDDPC_overview}
  \vspace{-2em}
\end{figure}

In this paper, we propose a novel \emph{Hybrid Data-Driven Predictive Control (HDDPC)} framework (Fig. \ref{fig:overview}) that simultaneously plan both contact schedules and continuous domain trajectories. Our framework leverages a data-driven reduced-order model based on Hankel Matrices. Building on prior work applying DDPC to locomotion \cite{li2024data}, we extend its application to include discrete step-to-step (S2S) dynamics. This unified framework for foot contact scheduling and trajectory planning integrates data from the discrete and continuous hybrid dynamics of locomotion, resulting in data-driven adaptation in a hybrid setting aimed at robust and dynamic locomotion (see Fig. \ref{fig:HDDPC_overview}). To demonstrate this method, we apply it to the lower-body exoskeleton Atalante, both in simulation and on hardware. We show that HDDPC enables stable walking at different speeds and effectively reject disturbances. The results indicate that a hybrid data-driven approach to exoskeleton locomotion can support the synthesis of robust and reactive walking gaits. 


\section{Models of Robotic Bipedal Locomotion}
\label{sec:models}
Consider a robotic system with generalized coordinates $q = \col(q_b, q_a) \in \mathcal{Q} \subset \mathbb{R}^n$, where ``$\col$'' indicates the column operator. The floating-base coordinates are $q_b \in SE(3)$, while the actuated degrees of freedom (DoFs) are $q_a \in \mathbb{R}^m$. The control input is $u \in \mathbb{R}^m$. The full-order state is represented as $x = \col(q, \dot{q}) \in T\mathcal{Q}$, where $T\mathcal{Q}$ denotes the tangent bundle of the configuration manifold.

\noindent \textit{\underline{Hybrid Dynamics:}} 
Legged locomotion is often modeled as a hybrid dynamical system composed of continuous domains representing continuous dynamics, and edges that capture changes in contact and discrete transitions such as impact events between domains\cite{westervelt2003hybrid}. 
Let $\mathcal{D}$ denote the domain and $\mathcal{S}$ the guard of the system. The hybrid system $\mathcal{H}$ can be described as:
\begin{numcases}{\mathcal{H}:}
\dot{x} = f(x) + g(x)\,u & $x \in \mathcal{D} \setminus \mathcal{S}$, \label{eq: continuous_dynamics}
\\
x^+ = \Delta(x^-) & $x^- \in \mathcal{S}$, \label{eq: discretecontrol}
\end{numcases}
where \eqref{eq: continuous_dynamics} represents the continuous full-order Lagrangian dynamics, while $\Delta: \mathcal{S} \rightarrow \mathcal{D}$ represents the discrete event when contact mode changes, and the superscripts ``$-$'' and ``$+$'' indicating the moments immediately before and after the discrete event, respectively. Typically, the events that trigger transitions between domains depend on the vertical position of the swing foot (e.g., impact occurs when its height equals the ground height) or the values of ground reaction forces (e.g., lift-off occurs when the normal force becomes zero).

Within each domain, the system dynamics can be derived from Euler-Lagrange equation:
\begin{align}
  D(q)\ddot{q} + H(q,\dot{q}) = B u + J_h(q)^TF \label{eq:dynamics} \\
  J_h(q) \ddot{q} + \dot{J}_h(q,\dot{q})\dot{q} = 0 \label{eq:hol_dynamics}
\end{align}
where $D(q):\mathcal{Q}\to \mathbb{R}^{n\times n}$ is the mass-inertia matrix, $H: T\mathcal{Q} \to \mathbb{R}^n$ contains the Coriolis and gravity terms, and $B\in\mathbb{R}^{n\times m}$ is the actuation matrix. The Jacobian of the holonomic constraint for the the contact is $J_h(q) \in \mathbb{R}^{h \times n}$ and the constraint wrench is $F \in \mathbb{R}^h$. During the single-support phase, there are $h=6$ for patched contact. 

\noindent \textit{\underline{Step-to-Step Dynamics:}} Step-to-Step (S2S) dynamics in legged locomotion are represented by a discrete-time hybrid system representation via the Poincaré return map, $P: \mathcal{S} \to \mathcal{S}$. This map converts the continuous hybrid dynamics into a discrete-time system by evaluating the state at successive points on the Poincaré section, resulting in $x_{k+1} = P(x_k)$ for $k = 0,1,\dots$. The discrete-time representation enables stability analysis of the overall system through the stability of the S2S dynamics. Specifically, the periodic orbit $\mathcal{O}$ is exponentially stable if the fixed point of the return map, $x^*$, is stable, as shown in Theorem 1 of \cite{morris2005restricted}.


\noindent \textit{\underline{Inverted Pendulum Models:}}
By constraining angular momentum to be constant (i.e. $\dot{\bm{L}}_{\CoM} = 0$), the Newton-Euler equation for the under-actuated floating-base coordinates simplifies to $ \coppos^{\{x,y\}} = \compos^{\{x,y\}} - \frac{\compos^z - \coppos^z}{\comacc^z+g}\comacc^{\{x,y\}} \label{eq:cop_com}$
where $\compos$ is center of mass (CoM) position, $\comacc$ is CoM acceleration. Additionally $\coppos^{\{x,y\}} = \frac{\sum_i f^z_i \bm{p}_i^{\{x,y\}}}{\sum_i \bm{f}_i^z}$ represents the Center of Pressure (CoP), with $i$ indexing the contact point $\bm{p}_i$ and $g$ is the gravity vector. Since the contact forces $f_i$ are unilateral ($f_i^z \geq 0$), CoP must lie within the convex hull of contact points ($\coppos \in \text{conv}\{\bm{p}_i\}$). By further constraining $\compos^z$ to be constant, the classic linear inverted pendulum model (LIP) is derived. LIP, along with the Zero Moment Point (ZMP) approach, is widely used for humanoid walking \cite{kajita2003biped, kajita2002realtime}. In this context, LIP is continuously actuated and can be interpreted as using its ZMP to approximate the ZMP of the full system with LIP dynamics embedded.

\noindent \textit{\underline{Hybrid-LIP:}}
To capture the hybrid nature of bipedal walking, \cite{xiong20223} introduced the Hybrid-Linear Inverted Pendulum (H-LIP) model. This model approximates the hybrid walking dynamics of an underactuated bipedal robot under the assumption of constant CoM height and stepping frequency, similar to the LIP model. The difference between the model is treated as a bounded error, which is used to design state-feedback stepping controllers that stabilize the horizontal CoM states within bounded invariant sets. The H-LIP dynamics are described as follows:
\begin{align*}
    \comacc^{\{x,y\}} = \frac{g}{\compos^z} \compos^{\{x,y\}}\text{ (SSP)} \quad \quad \quad \comacc = 0\text{ (DSP)}
\end{align*}
where the $\compos^z$ and domain duration ($T_{\textrm{SSP}}$, $T_{\textrm{DSP}}$) are constant. Assuming no velocity jump, we can combine the two domains with step size $\lambda$ equivalently as:
\begin{numcases}{\Delta_{\textrm{SSP}^- \rightarrow \textrm{SSP}^+}:} 
\comvel^+ = \comvel^- \nonumber\\ \compos^+ = \compos^- + \comvel^-T_{\textrm{DSP}}- \lambda.\nonumber \end{numcases}
H-LIP model S2S dynamics are given by $x_{k+1}^{\textrm{H-LIP}} = Ax_{k}^{\textrm{H-LIP}} + B\lambda_{k}$, with $x^{\textrm{H-LIP}} = [\compos,\comvel]^\top$, which can be used to approximate the Poincaré return map of the full order system.

\begin{figure}
  \centering
  \noindent\includegraphics[width=\linewidth]{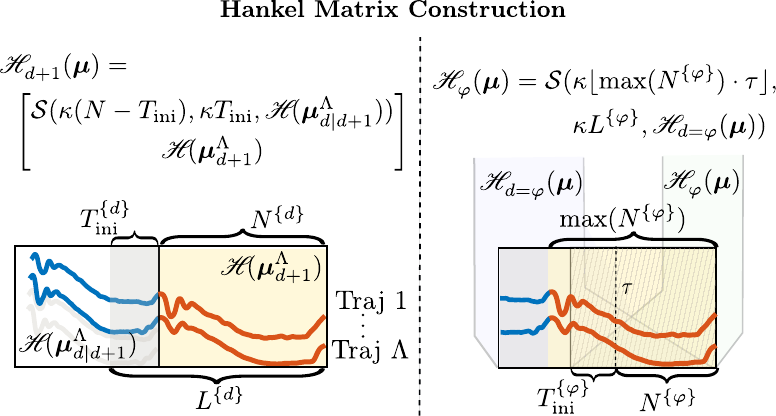}
  \caption{An illustration of \thm~construction for different domains. 
  }
  \label{fig:hm_construction}
  \vspace{-2em}
\end{figure}

\section{Hybrid Data-driven Model}
We introduce a hybrid data-driven model for full-order system dynamics (Fig. \ref{fig:hm_construction}), consisting of two components: a Hankel matrix $\HC{}{\cdot}$ for continuous dynamics and a second one $\HD{}{\cdot}$ for discrete step-to-step dynamics. 

\subsection{Trajectory Hankel matrix}
This subsection briefly reviews key results from behavioral systems theory and how \thm~is used as a data-driven model. We consider a state representation of a discrete-time LTI system as follows:
\begin{equation}
 \begin{aligned}
   & \theta(t+1) &&= A\,\theta(t) + B\,\mu(t) \\
   & \eta(t)     &&= M\,\theta(t) + D\,\mu(t),
 \end{aligned} \label{eq: LTI}
 \end{equation}
where \( \theta(t) \in \mathbb{R}^\beta \), \( \mu(t) \in \mathbb{R}^\kappa \), \( \eta(t) \in \mathbb{R}^\nu \) represent the state vector, control inputs, and outputs, respectively, at time \( t \in \mathbb{Z}_{\geq 0} := \{0,1,\dots \} \), and \( A \in \mathbb{R}^{\beta \times \beta} \), \( B \in \mathbb{R}^{\beta \times \kappa} \), \( M \in \mathbb{R}^{\nu \times \beta} \), \( D \in \mathbb{R}^{\nu \times \kappa} \) denote the unknown state matrices. In behavioral systems theory, a dynamical system is defined as a 3-tuple $(\mathbb{Z}_{\geq 0},\mathbb{W},\mathscr{B})$, where $\mathbb{W}$ is a signal space and $\mathscr{B} \in \mathbb{W}^{\mathbb{Z}_{\geq 0}}$ is the behavior. In contrast with classical systems theory with a particular parametric system representation such as that of \eqref{eq: LTI}, behavioral systems theory focuses on the subspace of the signal space where system trajectories live.

Let \(\bm{\mu} := \text{col}(\mu_0, \mu_1,\dots,\mu_{T-1})\) be an input trajectory with length \( T \in \mathbb{N} := \{1, 2, \dots\} \) applied in \(\mathscr{B}\) with the corresponding output trajectory \(\bm{\eta} := \text{col}(\eta_0, \eta_1,\dots,\eta_{T-1})\). Considering input sequences \( \bm{\mu}^{\Lambda} := \{\bm{\mu}^{(i)} \mid \forall i = 1, \dots, \Lambda\}\), where \(\Lambda\) indicates the number of datasets, we can construct the \thm~as follows:
\[
\Hankel{}{\bm{\mu}^\Lambda} := 
\begin{bmatrix}
\mu_0^{(1)} & \mu_0^{(2)} & \cdots & \mu_0^{(\Lambda)} \\
\mu_1^{(1)} & \mu_1^{(2)} & \cdots & \mu_1^{(\Lambda)} \\
\vdots & \vdots & \ddots & \vdots \\
\mu_{L-1}^{(1)} & \mu_{L-1}^{(2)} & \cdots & \mu_{L-1}^{(\Lambda)}
\end{bmatrix} \in \mathbb{R}^{\kappa L \times \Lambda},
\]
where \(\mu_j^{(i)}\) represents the \(j\)-th sample from the \(i\)-th trajectory or dataset. The \thm~with $\bm{\eta}^{\Lambda}$, denoted by $\Hankel{}{\bm{\eta}^\Lambda}$, can be constructed analogously.

The input signal $\bm{\mu}^\Lambda$ is persistently exciting of order L if $\Hankel{}{\bm{\mu}^\Lambda}$ is of full row rank \cite{van2020willems}. If a collection of  input-output (I-O) trajectory of the system, denoted $(\bm{\mu}^{\Lambda},\bm{\eta}^{\Lambda})$, is persistently exciting of order $L+\beta$, then by the Fundamental Lemma and it's extension to multiple dataset \cite{coulson2019data, Berberich_DDPC, willems2005note}, any trajectory of the LTI system can be constructed as a linear combination of the columns of the \thms. Consequently, any new trajectory pair of length $L$, $(\bm{\mu}_L,
\bm{\eta}_L)$ lies in the range space of the \thms, meaning it's a valid trajectory as long as there exists a vector $\gamma \in \mathbb{R}^\Lambda$ satisfying the following constraint:
\begin{align}
\begin{bmatrix}
\Hankel{}{\bm{\mu}^{\Lambda}} \\
\Hankel{}{\bm{\eta}^{\Lambda}}
\end{bmatrix} \gamma = \begin{bmatrix}
\bm{\mu}_L \\
\bm{\eta}_L
\end{bmatrix},
\label{eq:traj_constraint}
\end{align}
When using \thms~for planning, we partition them into two components: the initial condition estimation (past) and prediction portion (future) as follows: 
\begin{align}
\label{eq:hankel_partition}
\begin{bmatrix}
U_p \\
U_f
\end{bmatrix}
:= \Hankel{}{\bm{\mu}^{\Lambda}} \quad \quad
\begin{bmatrix}
Y_p \\
Y_f
\end{bmatrix}
:= \Hankel{}{\bm{\eta}^{\Lambda}},
\end{align}
where $U_p \in \mathbb{R}^{\kappa T\ini\times \Lambda}$ and $Y_p \in \mathbb{R}^{\nu T\ini\times \Lambda}$ are the portions for estimation with $T\ini$ length, while $U_f \in \mathbb{R}^{\kappa N\times \Lambda}$ and $Y_f \in \mathbb{R}^{\nu N\times \Lambda}$ are the portions for prediction. To streamline the partitioning, we define a selection operator $\Select(\cdot,\cdot,\cdot)$ as follows $\vphantom{U_p} U_p=\! \Select(0,\kappa T\ini, \Hankel{}{\bm{\mu}^{\Lambda}}), \;
 U_f=\! \Select(\kappa T\ini,\kappa N, \Hankel{}{\bm{\mu}^{\Lambda}})$. The first entry indicates the starting point of the selection, the second entry indicates the size of row block being selected. We can partition $\Hankel{}{\bm{\eta}^{\Lambda}}$ in a similar fashion with $\nu T\ini$ and $\nu N$. We can combine \eqref{eq:traj_constraint} and \eqref{eq:hankel_partition} in a predictive control framework to plan for a future trajectory with prediction horizon $N$ and $L=T_{\textnormal{ini}}+N$. 

\begin{figure}
  \centering
  \noindent\includegraphics[width=0.97 \linewidth]{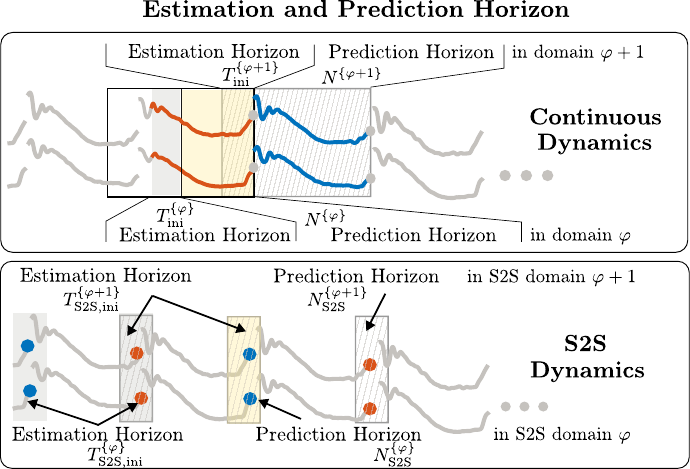}
  \caption{An illustration of the estimation and prediciton horizon of HDDPC}
  \label{fig:hm_horizon}
  \vspace{-2em}
\end{figure}

\subsection{Data-driven Model with Step-to-step Dynamics}
Inspired by the input-output structure of the H-LIP model in Section \ref{sec:models}, we define the input $\SSInput \in \R^3$ and the output $\SSOutput \in \R^2$ of the S2S date-driven model as follows:
\begin{align*}
    \bm{\SSInput} := \col(\lambda^{x,y}, T_{\text{step}}) \quad \quad \bm{\SSOutput} := (\compos^{x,y})^-,
\end{align*}
where $\lambda^{x,y}$ is the foot placement for the next step and $T_{\text{step}}$ is the desired step duration, and $(\compos^{x,y})^-$ are the pre-impact CoM states.
By using this I-O structure, the data-driven model captures the essential S2S dynamics that govern transitions in walking, allowing us to approximate the full system’s behavior through a step-wise dynamics evolution. We denote the S2S \thm~constructed using a dataset $\mu^{(i)}_{\textnormal{S2S}}$ that describe step-to-step transitions from domain $\mathcal{D}^d$ to $\mathcal{D}^{d+1}$ as $\HankD{\text{S2S}}{d}{\bm{\mu}^\Lambda_{\textnormal{S2S}}}$, where $d\in\mathbb{Z}_{\geq 0}$.

\subsection{Data-driven Model with Continuous Dynamics}
Drawing the inspiration from LIP model introduced in Section \ref{sec:models}, the input $\ContInput \in \R^2$ and the output $\ContOutput \in \R^2$ are chosen as follows:
\begin{align*}
    \ContInput := \coppos^{x,y} \quad \quad \ContOutput := \compos^{x,y}
\end{align*}

When considering multiple continuous domains, each domain is represented by its own \thm, capturing the system dynamics specific to that walking phase. Since we focus exclusively on position outputs, there are no discrete jumps in the CoM trajectory across domains in the global coordinate frame. However, each trajectory is expressed relative to the corresponding stance foot, resulting in a periodic oscillations in the $x$-direction and a sign flip in the $y$-direction as stance foot changes.
To account for transitions between domains, we apply a frame transformation ${\Phi(\cdot,\cdot)}$ to express CoM trajectory relative to the stance foot frame of the next domain. Specifically, when transitioning from domain $\mathcal{D}^d$ to $\mathcal{D}^{d+1}$, the transformation is given by $\bm{\mu}_{d |d+1}= \Phi(\bm{\mu}_{d},\lambda_{d}^{x,y})$, where $\lambda_{d}^{x,y}$ is the foot placement during the transition, measured relative to the stance frame in domain $\mathcal{D}^d$. Here $\bm{\mu}_{d}$ represents the CoM trajectory in domain $\mathcal{D}^d$ relative to its stance foot frame, while $\bm{\mu}_{d | d+1}$ denotes the same trajectory but relative to the stance foot frame in $\mathcal{D}^{d+1}$.
For a trajectory dataset $\bm{\mu}^\Lambda$ containing multiple domains, we can construct the \thm~$\HankD{}{{d+1}}{\bm{\mu}} \in \R^{\kappa L \times \Lambda}$ that describe the dynamics for domain $\mathcal{D}^{d+1}$ as follows: 
\begin{align}
\label{eq:hankel_hddpc}
   \HankD{}{{d+1}}{\bm{\mu}}
   =\begin{bmatrix} \Select(\kappa (N-T\ini),\kappa T\ini,
\Hankel{}{\BoldContInput_{d |d+1}^{\Lambda}}) \vspace{0.2em} \\ 
\Hankel{}{\BoldContInput_{d+1}^{\Lambda}}\end{bmatrix}, 
\end{align}
where $\BoldContInput_{{d}}^{\Lambda}$ denotes the trajectory data set captured in domain $d$ and $\Hankel{}{\BoldContInput_{d}^{\Lambda}},\Hankel{}{\BoldContInput_{d |d+1}^{\Lambda}} \in \R^{\kappa N \times \Lambda}$. Since all trajectories are transformed with respect to the stance foot frame in domain $\mathcal{D}^{d+1}$, continuity in the transition between adjacent domains is thereby guaranteed in \thm~construction. Similarly, $\HankD{}{{d+1}}{\bm{\eta}}$ can be constructed as illustrated in \eqref{eq:hankel_hddpc}.



\subsection{Hybrid Data-Driven Predictive Control}
Next, we present a planning problem over continuous trajectory for $K+1$ domains indexed by $d$, where $d\in\mathbb{Z}_{\geq 0}$ and $J$ step-wise dynamics indexed by $j$, where $j$ should be the corresponding transition between $d$-th and $d+1$-th domain. Each domain has its own continuous \thms~and each step-wise transition is associated with its corresponding S2S \thms. 

\noindent \textit{\underline{Prediction Horizon and Estimation Horizon}} The total planning horizon over the $K+1$ domains is defined to be $N_{\textnormal{tot}} = \sum_{d=\varphi}^{\varphi+K} N^{\{d\}}$, where $N\dom$ represents the prediction horizon for $d$-th domain. 
When planning over $K+1$ domains, for all the upcoming domains (i.e. $d > \varphi$), this prediction horizon is fixed (i.e.  $L\dom=T\ini\dom +N\dom, \; \forall d>\varphi$), which is the maximum possible prediction horizon determined by the construction. But for current domain $d=\varphi$, we are shrinking the prediction horizon as we move along the domain.




\noindent \textit{\underline{Data-driven Constraint}}
To construct the data-driven constraints similar to \eqref{eq:traj_constraint}, we select the appropriate portion of the \thms~and compose it to have the structure in \eqref{eq:hankel_hddpc}. 
Analogous to \eqref{eq:hankel_partition}, $T\ini$ is for determining the initial condition estimation. 
However, since for current domain the prediction horizon is changing, we have to partition the \thm~differently. Specifically, we need to select the appropriate $\kappa (T_\textnormal{ini}+N^{\{\varphi\}})$ rows from $\HankD{}{\varphi}{\bm{\mu}}$, starting from $\kappa\lfloor \max(N^{\{\varphi\}}) \cdot \tau\rfloor$-th row, where $\tau \in [0,1]$ indicates a phasing variable evolve from $0$ to $1$ on each step. Additionally, $\max(N^{\{\varphi\}})$ denotes the maximum prediction horizon for the current domain $\varphi$, which happens when $\tau=0$.

With a slight abuse of notation, we define $\HankD{}{{\varphi}}{\bm{\mu}}$ to be
\begin{align*}
       \HankD{}{{\varphi}}{\bm{\mu}}=
\Select(\kappa \lfloor \max(N^{\{\varphi\}})\cdot \tau \rfloor,\kappa L^{\{\varphi\}},\Hankel{{d=\varphi}}{\bm{\mu}}),
\end{align*}
where $\Hankel{{d=\varphi}}{\bm{\mu}}$ indicates the \thm~constructed analogous to \eqref{eq:hankel_hddpc} when $d=\varphi$, while $\HankD{}{{\varphi}}{\bm{\mu}}$ indicates the extracted portion from $\Hankel{{d=\varphi}}{\bm{\mu}}$ when $\tau$ evolves during the gait in current domain $\varphi$.

Since we are dealing with nonlinear systems, we added slack variable $\sigma\dom$ to account for noise, unknown requirement for persistently exciting order  and ensure numerical feasibility:
\begin{align}
\label{eq:hankel_withslack}
\small
    \begin{bmatrix}
\Hankel{{d}}{\bm{\mu}}\\
\Hankel{{d}}{\bm{\eta}}
\end{bmatrix} \gamma\dom +
\sigma\dom  =
\begin{bmatrix}
\bm{\mu}_{\textnormal{ini}}\dom \vspace{0.2em}\\
\bm{\mu}\dom \vspace{0.2em}\\
\bm{\eta}_{\textnormal{ini}}\dom\vspace{0.2em}\\
\bm{\eta}\dom
\end{bmatrix}, 
\end{align}
where $\bm{\mu}_{\textnormal{ini}}\dom$ and $\bm{\eta}_{\textnormal{ini}}\dom$ are the $T_\textnormal{ini}$ data points from the past trajectory data. The past trajectory data is either in the current domain only, or the portion of the previous domain can be collected in the past trajectory data set as the estimation horizon ,$T_\textnormal{ini}$, is fixed. Notably, the portion of the past trajectory data is subject to the frame transformation, $\Phi (\cdot, \cdot)$, in the case that the data from previous domain is employed in $\bm{\mu}_{\textnormal{ini}}\dom$ and $\bm{\eta}_{\textnormal{ini}}\dom$ construction.

We further note that the trajectory planning is over both discrete points $\{\bm{\mu}\dom,\bm{\eta}\dom\}$ and Bezier coefficient $\bm{\alpha}_{\text{com}^{x,y}}\dom$ so that B\'ezier function, $\text{Bez}(t_k\dom, \bm{\alpha}\dom,T\dom)$, and its derivative $\text{dBez}(t_k\dom, \bm{\alpha}\dom,T\dom)$, where $t_k\dom = T\dom \cdot \tau_k\dom$ and $\tau_k\dom = \{0,\delta_\tau\dom,\dots,1\}$, is used to obtain a smooth trajectory for low level controller. Note that $\delta_\tau$ is a sampling period that help us keep the same shape of \thm~across datasets with different duration for the same phase. 
Now we introduce our \textbf{Hybrid Data-Driven Predictive Controller (HDDPC)}:
\begin{align}
\label{eq:hybrid_ddpc}
& \underset{X} \min
& & \!\!\!\!\! \sum_{k=0}^{N_\textnormal{tot}} \left(\|\eta_k - r^\eta_{k}\|_Q^2 + \|\mu_k - r^\mu_{k}\|_R^2\right) + \psi_\gamma\|\gamma\|^2 + \psi_\sigma \|\sigma\|^2 \notag\\
& \,\,\, \text{s.t}
& & 
\!\!\!\!\!\! \begin{bmatrix}
\Hankel{d}{\bm{\mu}}\\
\Hankel{d}{\bm{\eta}}
\end{bmatrix} \gamma\dom +
\sigma\dom  =
\begin{bmatrix}
\bm{\mu}_{\textnormal{ini}}\dom \vspace{0.2em}\\
\bm{\mu}\dom \vspace{0.2em}\\
\bm{\eta}_{\textnormal{ini}}\dom\vspace{0.2em}\\
\bm{\eta}\dom
\end{bmatrix} \nonumber\\
&&&\forall d \in \{\varphi, \dots \varphi+K\}  \nonumber\\
&&&
\!\!\!\!\!\! \begin{bmatrix}
\HankD{S2S}{j}{\bm{\mu}}\\
\HankD{S2S}{j}{\bm{\eta}}
\end{bmatrix} \gamma_\textrm{S2S}^{\{j\}} +
\sigma_\textrm{S2S}^{\{j\}}  =
\begin{bmatrix}
\bm{\mu}^{{\{j\}}}_{\text{S2S,ini}}\vspace{0.2em}\\
\bm{\mu}^{\{j\}}_\textrm{S2S} \vspace{0.2em}\\
\bm{\eta}^{{\{j\}}}_{\text{S2S,ini}}\vspace{0.2em}\\
\bm{\eta}^{\{j\}}_\textrm{S2S}
\end{bmatrix} \nonumber\\
&&&\forall j \in \{\varphi,\dots,\varphi+K,\dots, \varphi + J\}, \; J\geq K  \nonumber\\
&&& \mu_k\dom \in \text{Support Polygon} \\
&&&  \eta_k\dom = \text{Bez}(t_k\dom, \bm{\alpha}_{\text{com}^{x,y}}\dom,T_{\textnormal{step}}^{\{d\}}) \nonumber \\
&&& \eta_{\text{S2S}}^{{\{j\}}} = \text{Bez}(T_{\textnormal{step}}^{\{j\}}, \bm{\alpha}_{\text{com}^{x,y}}^{\{j\}},T_{\textnormal{step}}^{\{j\}}) \nonumber\\
&&& \comvel(t_k\dom) = \text{dBez}(t_k\dom, \bm{\alpha}_{\text{com}^{x,y}}\dom,T_{\textnormal{step}}^{\{d\}}) \in [\comvel^{\text{min}},\comvel^{\text{max}}]  \nonumber \\
&&& \compos(t_0)  = \text{Bez}(t_0,\bm{\alpha}_{\text{com}^{x,y}}^{\{\varphi\}},T_{\textnormal{step}}^{\{\varphi\}}-t_0) \nonumber \\
&&& \comvel(t_0)  = \text{dBez}(t_0,\bm{\alpha}_{\text{com}^{x,y}}^{\{\varphi\}},T_{\textnormal{step}}^{\{\varphi\}}-t_0), \nonumber
\end{align}
where $Q$ and $R$ are positive definite matrices, $\psi_\gamma$ and $\psi_\sigma$ are positive weighting factors, 
and the decision variables $X\dom$, $X^{\{j\}}_{\text{S2S}}$, and $X$ are defined as
\begin{align*}
X\dom &= \col(\bm{\alpha}_{\text{com}^{x,y}}\dom, \bm{\mu}\dom, \bm{\eta}\dom, \bm{\gamma}\dom, \bm{\sigma}\dom) \\
X^{\{j\}}_{\text{S2S}} &= \col(\bm{\mu}^{\{j\}}_{\text{S2S}},\bm{\eta}^{\{j\}}_{\text{S2S}}, \bm{\gamma}^{\{j\}}_{\text{S2S}}, \bm{\sigma}^{\{j\}}_{\text{S2S}}) \\
X &= \col(X^{\{\varphi\}}, \dots, X^{\{\varphi+K\}}, X^{\{\varphi\}}_{\text{S2S}}, \dots, X^{\{\varphi+J\}}_{\text{S2S}}),
\end{align*}
respectively
and
the reference trajectory for $\mu$ and $\eta$ are denoted by $r^\mu$ and $r^\eta$ respectively.

\section{Layered Data-Driven Control Framework}

\begin{figure*}
  \centering
  \noindent\includegraphics[width=\linewidth]{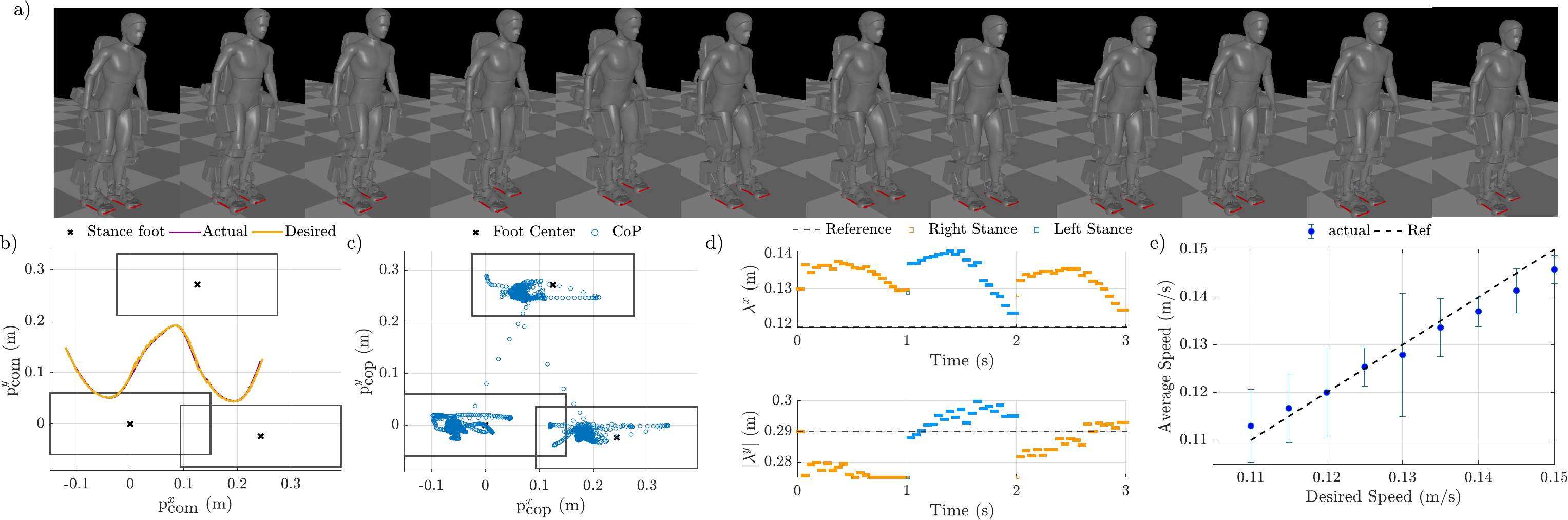}
  \caption{a) Gait tiles for the resulting trajectory b) Desired CoM trajectory from HDDPC planner and actual evolving CoM trajectory in simulation in global coordinate c) actual CoP in simulation d) planned foot placement location for three steps. e) the tracking performance for the HDDPC planner under different desired speed vs. actual realized average speed}
  \label{fig:hybrid_traj}
  \vspace{-0.85em}
\end{figure*}

 \begin{figure*}
  \centering
  \noindent\includegraphics[width=\linewidth]{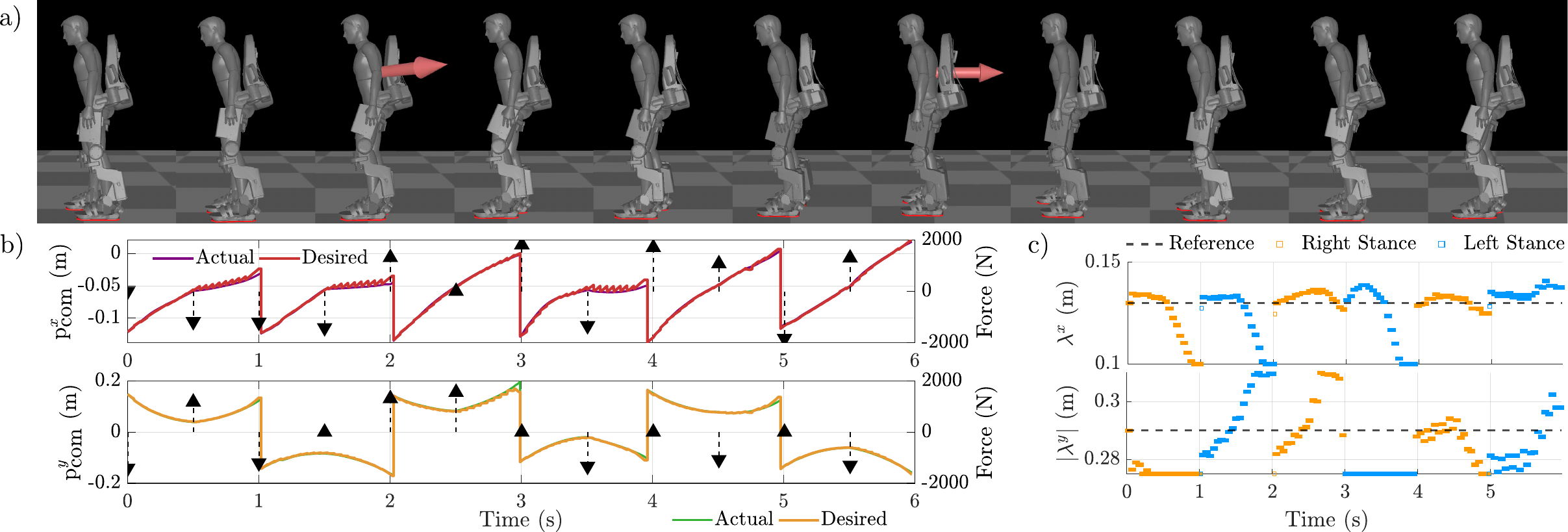}
  \caption{a) Gait tiles of simulated perturbation recovery b) CoM trajectory under random perturbation force. The time, direction, and magnitude of the perturbation is represented by the black arrows. The perturbation force is applied as a $10$ ms impulse with magnitude range between $1400$-$2000$ N c) The corresponding step location planned by the planner. The desired step size is indicated by the dashed line.} 
  \vspace{-1.5em}
  \label{fig:hybrid_ddpc_perturb}
\end{figure*}

In this section, we introduce the remaining components and details for practically implementing the layered control framework (Fig. \ref{fig:HDDPC_overview}) used to realize locomotion on the lower-body exoskeleton Atalante. We use a version of the model with the user height of 1.65 m and a total weight of 136 kg, which includes both the user's weight and the device.

\noindent \textit{\underline{Trajectory Hankel Matrix and Planner Parameters}}: Trajectory Hankel matrices are constructed from data collected over five gait cycles (10 steps), with step lengths varying between 0.11 m and 0.15 m and step durations ranging from 0.9 s to 1.1 s. For the continuous dynamics Hankel matrix, data is sampled at 50 evenly spaced points across the gait cycle (i.e., sampling period $\delta_\tau$ is $0.02$ for normalized step time). We set the estimation horizon that is used to determine system evolution for the trajectory Hankel matrix as $T_{\text{ini}} = 4$. The same data sequence is used to construct the S2S trajectory Hankel matrix. Specifically, we maintain one Hankel matrix for left stance, one for right stance, and two S2S matrices representing transitions from right-to-left (R2L) and left-to-right (L2R) dynamics. In total, we plan for trajectory over three domains (e.g. a Left-Right-Left sequence). In order to ensure a faster planning rate, we select a different $\delta_\tau$ for the second and third domain ($\delta_\tau = 0.08$) to reduce the number of decision variables required in \eqref{eq:hybrid_ddpc}. 

The planning problem is solved using IPOPT \cite{wachter2006implementation} with its C++ interface and HSL MA97 solver at $20$-$40$ Hz depending on the planning horizon and application scenario. We terminate the planner at a pre-specified maximum wall clock time and employ the feasible trajectories only (i.e. with constraint satification under tolerance). The planned trajectory, specifically $\{\bm{\alpha}_{\text{com}}\dom,\bm{\eta}_{\textnormal{S2S}}^{\{j\}},t\dom_0\}$ is sent to low-level controller.  When the planner is deployed, we do not update the step duration for the current domain. When an impact occurs, the step duration planned for the next domain will be used to evaluate the phasing variable.

\noindent \textit{\underline{Output Synthesis}}:
The desired walking behavior is encoded by the task space output $\mathbf{y} = \mathbf{y}_{\text{act}} - \mathbf{y}_{\text{des}}$, where $\mathbf{y}^{\text{act}} \in \R^{12}$ and $\mathbf{y}^{\text{des}} \in \R^{12}$ and are chosen to be the following
\begin{align*}
    \mathbf{y}&^{\text{act}} =  \begin{bmatrix}
    \composst^{x,y,z}(q) &  \phi^{x,y,z}_{\textrm{pelv}}(q) & \pos_{\textrm{sw}}^{x,y,z}(q) & \phi^{x,y,z}_{\textrm{sw}}(q)
    \end{bmatrix}\\
    \mathbf{y}&^{\text{des}} = \begin{bmatrix}
    \composst^{x,y}(\bm{\alpha}_{\text{com}^{x,y}}) & \composst^{z}(\bm{\alpha}) & \\ \phi^{x,y,z}_{\textrm{pelv}}(\bm{\alpha}) & \pos_{\textrm{sw}}^{x,y,z}(\bm{\alpha},\lambda^{x,y}) & \phi^{x,y,z}_{\textrm{sw}}(\bm{\alpha})
    \end{bmatrix}.
\end{align*}
The desired CoM position, foot placement, and step duration that determine the phasing variable is generated from the hybrid DDPC planner, and the other desired components are taken as B\'ezier polynomials with the coefficient matrix of $\bm{\alpha}$. More specifically, the coefficients of pelvis orientation $\phi_{\textrm{pelv}}$, swing foot orientation $\phi_{\textrm{sw}}$, $z$-height of CoM, and swing foot trajectory are fixed. The swing foot $x,y$ trajectories are determined by B\'ezier polynomials connecting the swing foot position at the beginning of the domain (i.e., post-impact state) and the desire foot targets, i.e., $\pos_{\textrm{sw}}^{x,y}(\tau) = (1-\beta(\tau))\,\pos_{\textrm{sw}}(q^+) + \beta(\tau)\,\lambda^{x,y}$, where $\beta$ is a phase-based weighting function.

\noindent \textit{\underline{Whole-body Controller}}: We apply a task-space based QP controller solved via OSQP \cite{osqp} with a maximum iteration of 200 running at 1kHz, formulated as following
\begin{align}\label{eq:wbc}
& \underset{(\ddot{q},u,F) \in R^{n+m+h}} \min
& & \|\ddot{\mathbf{y}}_{\text{act}} - \ddot{\mathbf{y}}^{fb}\|_Q^2 \\
& \text{subject to}
& & \eqref{eq:dynamics},\eqref{eq:hol_dynamics} \tag{dynamics}\\
& & & A^{GRF} F \leq b^{GRF} \tag{contact} \\
& & & u_{\text{min}} \leq u \leq u_{\text{max}}\tag{torque limit}
\end{align}
where $\ddot{\mathbf{y}}^{fb} = -\mathcal{K}_p\mathbf{y} - \mathcal{K}_d \dot{\mathbf{y}}$, $A^{GRF}$ describes friction cone constraint and zmp constraint for each contact frame.


\section{Experimental Validations}

\label{sec: result} In this section, we present the numerical simulation in MuJoCo \cite{todorov2012mujoco} and hardware experiment results to validate the effectiveness of our proposed framework. The experiment video can be found in \cite{video}.

\begin{figure*}
  \centering
  \noindent\includegraphics[width=\linewidth]{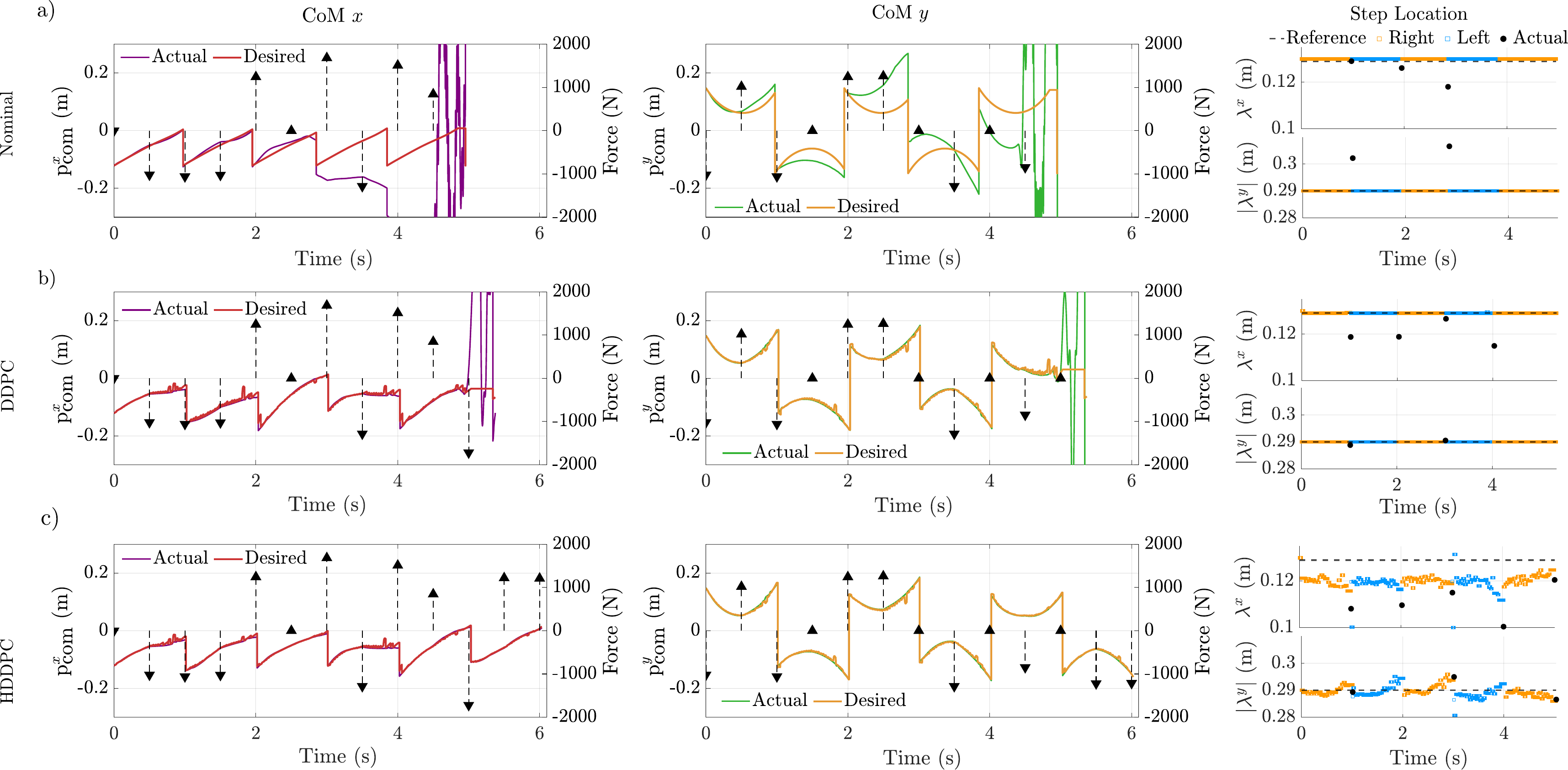}
  \caption{Perturbation Recovery Comparison: a) Nominal: The controller follows a fixed reference trajectory with a predetermined step size and step duration. b) DDPC: Functionally equivalent to HDDPC but with a fixed contact schedule. The upper and lower bounds of the step size and the step duration are constrained to match those used in the nominal reference trajectory. c) HDDPC: The proposed control framework.}
\vspace{-0.5em}
  \label{fig:perturn_sim_comp}
\end{figure*}

\noindent \textit{\underline{Tracking Performance (Simulation)}}: We first evaluate the planner's capability to realize stable walking. An example gait tiles from the MuJoCo simulation is shown in Fig. \ref{fig:hybrid_traj}a. An example planned and realized CoM trajectory, actual CoP position is shown in Figs. \ref{fig:hybrid_traj}b and \ref{fig:hybrid_traj}c, respectively. The planned step location are shown in Fig. \ref{fig:hybrid_traj}d over three steps. Additionally, we assess the tracking performance of the hybrid DDPC controller under different forward velocities, with the mean and standard deviation of the realized speed over a 10-second window plotted in Fig. \ref{fig:hybrid_traj}e. As the desired speed increases, tracking error also increases, but the realized walking remains stable.

\begin{figure*}
  \centering
\noindent\includegraphics[width=\linewidth]{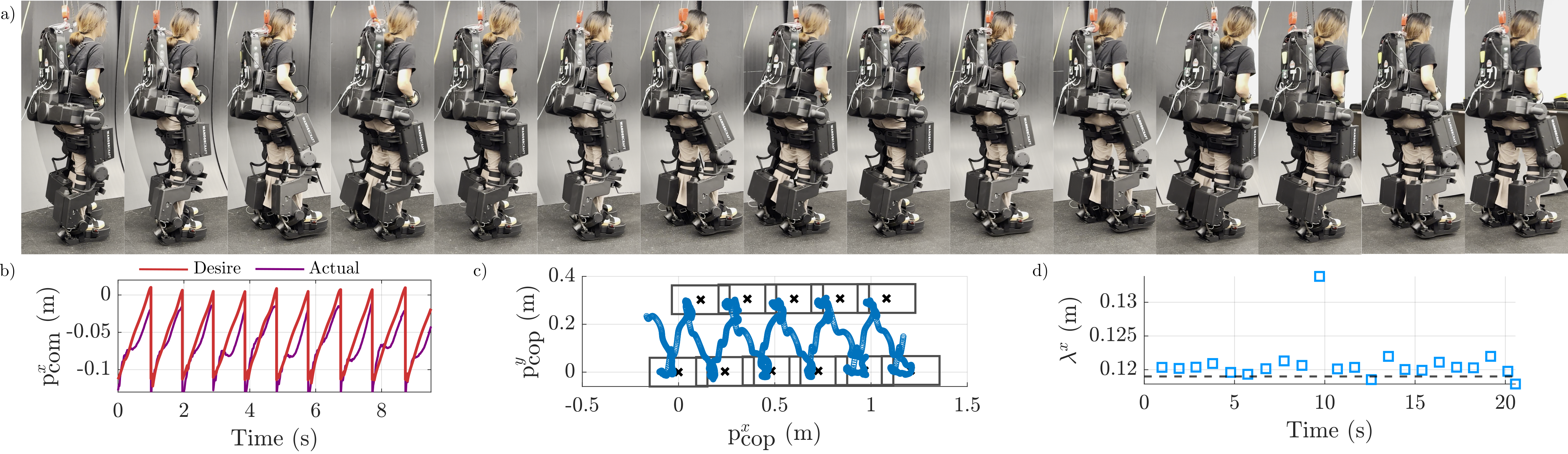}
  \caption{a) Gait tiles of walking on hardware together with b) CoM trajectory, c) CoP trajectory d) Planned foot placement}
  \label{fig:hardware}
  \vspace{-1.5em}
\end{figure*}

\noindent \textit{\underline{Perturbation Recovery (Simulation)}}: We further evaluate the framework's robustness by introducing external perturbations (see Fig. \ref{fig:hybrid_ddpc_perturb}a). Impulses were applied to the torso at intervals of $0.5$ s, lasting for 10 ms, with random perturbation directions selected from 45-degree intervals in the horizontal plane (i.e. 0°, 45°, etc.) and varying maximum magnitudes ($1400-2000N$). Figure \ref{fig:hybrid_ddpc_perturb}b illustrates the continuous CoM trajectory, highlighting the moments when perturbations occurred. The top-down view of the step pattern in Fig. \ref{fig:hybrid_ddpc_perturb}c shows the controller’s adjustments in response to each disturbance. 

We compared our proposed framework with two baselines. The first is a nominal approach that tracks a fixed reference trajectory with a predetermined step size and step duration. While a desired step duration is set, transitions to the next domain are enforced based on a combination of a minimum phasing variable threshold and impact events detected by the force sensor, a mechanism shared across all three comparison cases discussed here. Additionally, we include another baseline which we referred to as DDPC. We argue that this is conceptually similar to our previous DDPC results \cite{li2024data} but implemented differently. This is essentially an HDDPC variant with a fixed contact schedule, where the lower and upper bounds are set to match those of the nominal approach. This setup essentially allows for replanning only over the CoM trajectory. Both DDPC and HDDPC use the same set of hyperparameters but with the only difference being the restricted contact schedule bound. 

We applied the same perturbation sequence by fixing the random seed number. The desired step duration was 1 s, but the nominal controller impacted earlier and failed to maintain the desired step size very well. As shown in Fig. \ref{fig:perturn_sim_comp}a, the robot began accumulating significant tracking errors around 3 s and failed shortly thereafter. In contrast, DDPC (Fig. \ref{fig:perturn_sim_comp}b) maintained system stability until 5 s before eventually failing, while demonstrating better adherence to the desired step duration and step size compared to the nominal approach. Meanwhile, HDDPC (Fig. \ref{fig:perturn_sim_comp}c) successfully rejected all disturbances.

\noindent \textit{\underline{Hardware Results}}: In hardware experiments (see Fig. \ref{fig:hardware}a), we first collected the data set from the exo when subject is in it. During the data collection, we varied the foot step length for each gait to enable the foot step adjustment capability in HDDPC.
Based on the collected data set, HDDPC planner was deployed to synthesize CoM motion and foot placement in the $x$ direction for the exoskeleton, running on an external PC (Intel i9-14900K CPU), which communicated with exo through a UDP network. The HDDPC planner performed trajectory replanning of the desired CoM trajectory and foot placement at the beginning of each domain. While the model parameters in the low-level controller did not incorporate detailed subject-specific information, the HDDPC implementation with the collected data set successfully demonstrated stable locomotion on exo with human subject without compromising stability. The detailed tracking performance is illustrated in Fig. \ref{fig:hardware}. The evolution of CoM trajectories effectively tracked the desired CoM trajectories (see. Fig. \ref{fig:hardware}b), while the CoP was successfully regulated within each stance foot as described in Fig. \ref{fig:hardware}c. Figure \ref{fig:hardware}d further illustrates that the HDDPC actively modulated foot step length to track the desired trajectories while maintaining locomotion stability. Within this capacity of HDDPC framework, the exo demonstrated the stable bipedal locomotion.





\section{CONCLUSIONS}
We presented a novel HDDPC framework that integrates contact scheduling with continuous domain trajectory planning for lower-body exoskeletons. Through both simulation and hardware experiments, the HDDPC framework demonstrated its capability to achieve stable and adaptive walking. In simulations, the integration of S2S dynamics and continuous domain trajectory enhanced the exoskeleton's reactive capabilities, enabling effective recovery from external disturbances. The hardware results confirmed the effectiveness of the HDDPC controller. Future work will focus on running the planner faster online, enhancing the framework's ability to adapt to time-varying perturbations by updating the Hankel matrix online, and demonstrating robustness across subject-to-subject variability. Additionally, we aim to extend application to more complex scenarios, such as stair climbing and other challenging setups in dynamic environments.

\bibliographystyle{IEEEtran}
\balance
\bibliography{IEEEabrv, References}


\end{document}